\documentclass{article} %
\usepackage{mathpazo}
\usepackage[margin=1in]{geometry}
\usepackage[numbers, compress, sort]{natbib}

\PassOptionsToPackage{newfloat,frozencache,cachedir=./minted}{minted}

\usepackage{booktabs,multirow,xcolor,colortbl,pifont}
\usepackage{siunitx}
\usepackage{graphicx}
\usepackage{stmaryrd}
\usepackage{algorithm}
\usepackage{algorithmic}
\usepackage{mathtools}
\usepackage{extarrows}

\usepackage{amsmath,amsfonts,bm}

\def\eqref#1{equation~\ref{#1}}

\def\1{\bm{1}}

\DeclareMathAlphabet{\mathsfit}{\encodingdefault}{\sfdefault}{m}{sl}
\SetMathAlphabet{\mathsfit}{bold}{\encodingdefault}{\sfdefault}{bx}{n}

\usepackage{hyperref}
\usepackage{url}
\usepackage{xspace}
\usepackage{subcaption}
\usepackage[capitalize,noabbrev]{cleveref}

\AddToHook{cmd/appendix/before}{%
    \crefalias{section}{appendix}%
    \crefalias{subsection}{appendix}
}

\usepackage{stmaryrd}          %
\usepackage{mathpartir}        %

\newcommand{\den}[1]{\llbracket #1 \rrbracket}

\newcommand{\bnfdef}{\;\;::=\;\;}
\newcommand{\bnfalt}{\;\mid\;}
\newcommand{\kw}[1]{\mathbf{#1}}

\newcommand{\doubleplus}{\mathbin{+\!\!+}}

\usepackage{amsthm}

\usepackage[most]{tcolorbox}
\tcbuselibrary{listings,minted,skins,breakable}
\usepackage[newfloat,frozencache,cachedir=./minted]{minted} %

\setminted{
  breaklines,
  breakanywhere,         %
  autogobble,            %
  tabsize=2,
  fontsize=\footnotesize,
  encoding=utf8,
  texcomments=false,
  linenos,
  numbersep=6pt,
  breaksymbol=\raisebox{0.5ex}{\tiny\ensuremath{\hookrightarrow}},
}

\tcbset{
  promptbox/.style={
    enhanced,
    breakable,
    sharp corners,
    boxrule=0.4pt,
    colback=black!1,
    colframe=black!15,
    coltitle=black,
    left=6pt,right=6pt,top=6pt,bottom=6pt,
    attach boxed title to top left={yshift=-2pt, xshift=2pt},
    boxed title style={
      size=small,
      colback=black!3,
      colframe=black!15,
      sharp corners
    },
  }
}

\tcbset{
  lang/.style      = { minted language=#1 },               %
  pygstyle/.style  = { minted options={style=#1} },        %
  nolinenos/.style = { minted options={linenos=false} },   %
  tinytext/.style  = { minted options={fontsize=\scriptsize} },
}

\newtcblisting{promptblock}[2][]{%
  promptbox,
  listing engine=minted,
  minted language=markdown, %
  minted options={},
  listing only,
  title={#2},
  #1
}

\newtcbinputlisting{\includeprompt}[3][]{%
  promptbox,
  listing engine=minted,
  minted language=markdown, %
  minted options={},
  listing only,
  title={#2},
  listing file={#3},
  #1
}

\title{Experience-Guided Reasoning with\\ Contextual LLM Programs}
\title{Experience-Guided Reasoning by Adapting LLM Programs}
\title{Adaptive Reasoning Through Inference-Time Strategy Generation}
\title{From Experience to Strategy: Learning Adaptive Reasoning Programs}
\title{Dynamic Agent Adaptation\\for Experience-Guided Reasoning}
\title{Learning to Generate Inference-Time Strategies from Experience}

\title{Inference-Time Strategy Adaptation for Experience-Guided Reasoning}
\title{Adapting the Entire Reasoning Process to Inference-Time Experience}
\title{Generating Strategies at Inference-Time from Experience}

\title{Dynamic Strategy Adaptation\\for Experience-Guided Reasoning}
\title{Experience-Guided Adaptation of\\Inference-Time Reasoning Strategies}

\author{%
    Adam Stein\textsuperscript{1}\thanks{Work done during an internship at AWS AI. Correspondence to: Matthew Trager \texttt{<mttrager@amazon.com>}.}, Matthew Trager\textsuperscript{2}, Benjamin Bowman\textsuperscript{2}, Michael Kleinman\textsuperscript{2},\\
    Aditya Chattopadhyay\textsuperscript{2}, Wei Xia\textsuperscript{2}, Stefano Soatto\textsuperscript{2} \and
    \textsuperscript{1}University of Pennsylvania, \textsuperscript{2}AWS AI 
}

\newcommand{\ourmethod}{Experience-Guided Reasoner\xspace}
\newcommand{\ourabbrev}{\textsc{EGuR}\xspace}
\newcommand{\proc}[3]{#2 \xLongrightarrow{#1} #3}
\newcommand{\guide}{Guide\xspace}
\newcommand{\updater}{Consolidator\xspace}

\begin{document}

\date{}
\maketitle

\begin{abstract}
Enabling agentic AI systems to adapt their problem-solving approaches based on post-training interactions remains a fundamental challenge. While systems that update and maintain a memory at inference time have been proposed, existing designs only steer the system by modifying textual input to a language model or agent, which means that they cannot change sampling parameters, remove tools, modify system prompts, or switch between agentic and workflow paradigms. On the other hand, systems that adapt more flexibly require offline optimization and remain static once deployed. We present \ourmethod (\ourabbrev), which generates tailored \textit{strategies}---complete computational procedures involving LLM calls, tools, sampling parameters, and control logic---dynamically at inference time based on accumulated experience. We achieve this using an LLM-based \textit{meta-strategy}---a strategy that outputs strategies---enabling adaptation of all strategy components (prompts, sampling parameters, tool configurations, and control logic). \ourabbrev operates through two components: a \textbf{\guide} generates multiple candidate strategies conditioned on the current problem and structured memory of past experiences, while a \textbf{\updater} integrates execution feedback to improve future strategy generation. This produces complete, ready-to-run strategies optimized for each problem, which can be cached, retrieved, and executed as needed without wasting resources. Across five challenging benchmarks (AIME 2025, 3-SAT, and three Big Bench Extra Hard tasks), \ourabbrev achieves up to 14\% accuracy improvements over the strongest baselines while reducing computational costs by up to 111×, with both metrics improving as the system gains experience.
 \end{abstract}

\section{Introduction}

Modern AI systems employ sophisticated \textit{strategies}---complex computational procedures involving LLM calls, tools, and control logic, often implemented using frameworks like SGLang \citep{sglang} or DSPy \citep{dspy}---to tackle challenging reasoning tasks. For example, systems solving International Mathematical Olympiad (IMO) problems dynamically decompose problems into subgoals, verify solutions through parallel sampling, and iteratively self-correct \citep{huang2025gemini, comanici2025gemini}. Similarly, code generation agents switch between different prompting approaches, adjust sampling temperatures, and selectively use tools based on problem complexity \citep{textgrad, dspy}.

While these strategies are often designed to be general-purpose and highly adaptive to inputs---particularly agents, which consist of loops of LLM and tool execution---they generally remain \textit{static} at inference time, unable to learn from experience. For example, without persistent memory, an AI system for mathematical reasoning will apply the same expensive multi-step decomposition every time it encounters a problem it has solved in the past (potentially even failing to solve it again), and will repeatedly make identical mistakes on problems it has failed to solve before.

To address this problem, different methods for adapting AI systems have been proposed. Approaches such as Dynamic Cheatsheet \citep{dynamic-cheat} and Buffer of Thoughts \citep{bot} maintain memory across problems, but this memory is used only for \textit{textual steering} of an LLM or agent. As such, they cannot change sampling parameters, add or remove tools, modify control logic, or cache successful computational procedures for direct reuse. On the other hand, offline methods like ADAS \citep{adas} can adapt strategies much more flexibly, but require expensive training phases and remain static once deployed, unable to continue learning from new experiences. 
The core challenge is to \textbf{adapt the computational structure of strategies from experience, to continually improve both accuracy and efficiency}.

To address this challenge, we present \ourmethod (\ourabbrev), a system designed to generate strategies dynamically at inference time. Our approach uses an LLM-based \textit{meta-strategy}---a strategy that outputs strategies---rather than steering existing strategies at runtime, it proposes new strategies. This enables generation of complete, tailored computational procedures for each problem that can be easily conditioned using a memory store and enables simple and effective caching of successful strategies for reuse. \ourabbrev operates through two main components: a \textbf{\guide} that generates multiple candidate strategies conditioned on the current problem and structured memory of past experiences, and a \textbf{\updater} that processes execution outcomes to improve future strategy generation. By generating and comparing multiple strategies per problem, the system learns the relative effectiveness of different strategies, leading to continual improvements in both accuracy and computational efficiency.

Our contributions are:
\begin{itemize}
    \item We introduce a system that generates strategies dynamically at inference time based on accumulated experience, enabling adaptation of all strategy components including prompts, sampling parameters, tools, and control logic (\cref{sec:method}).

    \item We formalize strategies as compositions of stateful \textit{processes} which map inputs to outputs while also updating a state. This provides a unified representation for diverse reasoning paradigms (workflows, agents, tool-use systems) that supports compositional cost tracking and execution tracing (\cref{sec:representation}).

    \item We demonstrate that generating multiple strategies per problem and comparing their effectiveness enables continual improvement in both accuracy and efficiency (\cref{fig:claude-ablation}).

    \item Across five challenging benchmarks (AIME 2025, 3-SAT, and three Big Bench Extra Hard tasks), \ourabbrev achieves up to 14\% accuracy improvements (\ourabbrev vs. Mem0 on 3-SAT) while reducing computational costs by up to 111x (\ourabbrev vs. DC on Object Counting), with both metrics improving as the system gains experience (\cref{fig:holdout}).
\end{itemize}
\section{Strategies For AI Systems}

In this section, we formalize strategies as compositions of stateful processes---functions taking inputs and a state and producing an output and an updated state---providing a unified way to describe any strategy in terms of its components. This yields a natural taxonomy of common strategies and enables us to describe feedback mechanisms as state-modification methods.

\begin{figure}
    \centering
    \includegraphics{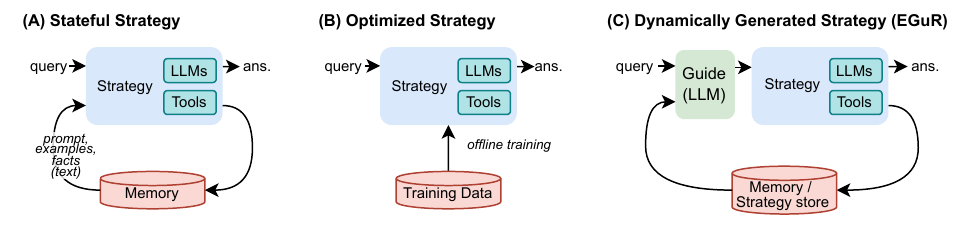}
    \caption{Comparison of existing experience-based adaptation methods (A and B) to our approach in (C) which dynamically produces a compiled strategy based on the current query and memory at inference time. Methods which augment existing strategies with state, such as Dynamic Cheatsheet\citep{dynamic-cheat} and Mem0\citep{mem0}, are depicted in (A) and methods such as ADAS \citep{adas} and OPTO \citep{cheng2024trace} which optimize strategies offline are shown in (B). Our method shown in (C) uses a state (which can contain useful compiled strategies) during inference to guide the system in producing effective strategies for each query, unlike existing methods which cannot adapt their strategies per-query.}
    \label{fig:overview}
\end{figure}

\subsection{Representing Strategies}
\label{sec:representation}

We use the term \textit{strategy} for a specification of the computational procedure that an AI system applies to inputs to produce outputs, capturing LLM inference calls, tool invocations, and control flow decisions. For example, the Chain-of-Thought (CoT) strategy involves a single LLM call with specific prompting, while CodeAct iteratively calls an LLM and executes generated code until a termination condition is met.

An adaptive strategy engages in a sequence of interactions, or episodes, and updates its internal state based on past outcomes.
To enable adaptive strategies, we need a representation that explicitly captures both what can be adapted and how state evolves across episodes. Existing frameworks like DSPy \citep{dspy} and SGLang \citep{sglang} implement strategies as Python code without explicitly marking adaptable components. TextGrad \citep{textgrad} and Trace \citep{cheng2024trace} mark components for optimization, but are designed for offline training and lack explicit state management for continual learning during inference. We introduce a compositional formalism that unifies adaptable components with state management, enabling meta-strategies to generate and refine strategies from accumulated experience.

\begin{figure}[t]
\centering
\begin{subfigure}[b]{0.68\textwidth}
\small
\vspace{0pt}
\[
\begin{array}{rcll}
S &::=& \mathbf{base}_P & \text{base process}\\
  &\mid& S_1;\; S_2 & \text{sequential composition where}\\
  &&& S_1: \proc{\Sigma_1}{A}{C} \text{ and } S_2: \proc{\Sigma_2}{C}{B}\\
  &&& \text{and } \Sigma_1, \Sigma_2 \text{ are views of }\Sigma\\\\
  &\mid& S_1 \parallel S_2 & \text{parallel composition where}\\
  &&& S_1: \proc{\Sigma_1}{A}{B_1} \text{ and } S_2: \proc{\Sigma_2}{A}{B_2} \text{ and}\\
  &&& B=(B_1, B_2) \text{ and } \Sigma_1, \Sigma_2 \text{ are disjoint views of }\Sigma\\\\
  &\mid& \mathbf{if}\ S_1\ \mathbf{then}\ S_2\ \mathbf{else}\ S_3 & \text{conditional where } S_1: \proc{\Sigma_1}{A}{\text{Bool}},\\
  &&& S_2: \proc{\Sigma_2}{A}{B}, \text{ and } S_3: \proc{\Sigma_3}{A}{B}\\
  &&& \text{and } \Sigma_1, \Sigma_2, \Sigma_3 \text{ are views of }\Sigma\\\\
  &\mid& \mathbf{recfun}\ f:\, S_1 & \text{recursive definition of process } f \text{ where}\\
  &&& S_1: \proc{\Sigma_1}{A}{B} \text{ which can internally use } f\\
  &&& \text{and } \Sigma_1 \text{ is a view of } \Sigma.
\end{array}
\]
\caption{Grammar for strategies where $S:\proc{\Sigma}{A}{B}$.}
\label{fig:grammar}
\end{subfigure}%
\hfill
\begin{subfigure}[b]{0.28\textwidth}
\vspace{0pt}
\small
\begin{center}
\begin{tabular}{@{}r@{\hspace{0.5em}}l@{}}
\textcolor{gray}{\footnotesize 1} & $\mathbf{recfun}\ \text{CodeAct}:$ \\[0.2em]
\textcolor{gray}{\footnotesize 2} & $\quad \text{CallLLM};\;$ \\[0.2em]
\textcolor{gray}{\footnotesize 3} & $\quad \mathbf{if}\ \text{ContainsCode}$ \\[0.2em]
\textcolor{gray}{\footnotesize 4} & $\quad \mathbf{then}\ (\text{ExecCode} ;\; \text{CodeAct})$ \\[0.2em]
\textcolor{gray}{\footnotesize 5} & $\quad \mathbf{else}\ \mathbf{return}$
\end{tabular}
\end{center}
\caption{Example: CodeAct Strategy}
\label{fig:codeact}
\end{subfigure}
\caption{High-level grammar for strategies as compositions of base processes (left) with an example of how CodeAct is represented in this grammar (right). A further formalization is provided in \cref{app:strategies} where we define how to additionally construct processes from pure functions as well as processes for explicitly accessing and updating the state, and some useful base processes such as $\mathbf{return}$.}
\label{fig:tiny-grammar}
\end{figure}

We model strategies as compositions of stateful \textit{processes}. A process is a function that takes an input and a state (e.g., conversation history, sampling parameters), produces an output, and updates the state. This explicit state management clarifies what components can be adapted (prompts, temperature, available tools).
Formally, we write a process type as $\proc{\Sigma}{A}{B}: (A \times \Sigma) \to (B \times \Sigma)$ where $A$ and $B$ are input and output domains and $\Sigma$ is the state space. Strategies are built compositionally using the grammar shown in Figure~\ref{fig:tiny-grammar}. Sequential composition ($S_1;\; S_2$) chains processes, parallel composition ($S_1 \parallel S_2$) executes concurrently, conditionals enable branching, and recursion enables iteration.

For instance, CodeAct (shown in Figure~\ref{fig:codeact}) composes an LLM call with a code execution tool recursively, iterating until the LLM output contains no code. The LLM call process maintains conversation history in its state, while the code execution process maintains the execution environment.

This compositional structure is key to our approach. By explicitly representing all components (prompts, sampling parameters, tools, control flow) and their state dependencies, we enable a meta-strategy to generate new strategies by composing different components. For instance, our meta-strategy can adapt CodeAct by changing the LLM's prompt, adjusting temperature, or removing the code interpreter entirely for non-algorithmic problems. This framework also allows us to formalize important properties of strategies: computational cost and intermediate traces (Appendix~\ref{app:cost-def}), formal semantics (Appendix~\ref{app:semantics}), and structural characteristics (discussed in the next section).

\subsection{The Strategy Landscape}

\begin{figure}[t]
    \centering
    \includegraphics[width=\linewidth]{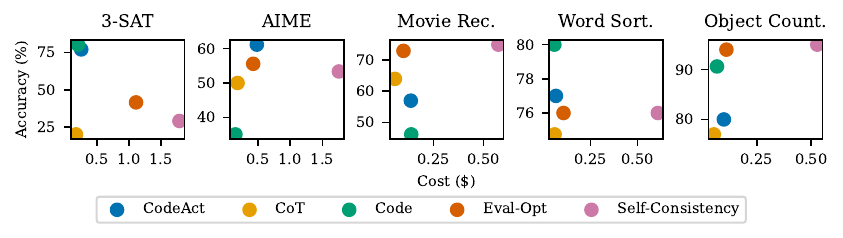}
    \caption{Comparison of strategy performance across tasks for Claude 3.7 Sonnet. Strategies closer to the top-left corner are best for the task in terms of accuracy and cost. The optimal strategy differs significantly across tasks. For example, Code
    excels on 3-SAT and Word Sorting but performs poorly on Movie Recommendation and AIME. Descriptions of the above strategies are provided in \cref{app:common-strats}.
   }
    \label{fig:strategy-diffs}
\end{figure}

We organize strategies into several categories based on their structure. A strategy is \textit{dynamic} if it contains conditionals, meaning the control flow can change at runtime based on the input. A dynamic strategy that is recursive (uses \textbf{recfun} as shown in \cref{fig:grammar}) is called an \textit{agent}. Strategies that do not use recursion are called \textit{workflows} since their runtime behavior is less dynamic, and a workflow that is not dynamic is called a \textit{pipeline}. Strategies can also use parallelization to solve several tasks concurrently or to produce multiple responses at once, which can improve accuracy and reliability \citep{wang2023selfconsistency}. Table~\ref{tab:strategies} shows a breakdown of five commonly used strategies based on their use of parallel compute, their dynamic or agentic nature, and their tool utilization. A description of the five strategies is in \cref{app:common-strats}. Our classification differs from common definitions that do not classify Eval-Opt as agentic and lack a clear way to determine when a strategy is agentic \citep{anthropic2024building}.

\newcommand{\cmark}{\ding{51}}
\newcommand{\xmark}{\ding{55}}
\newcommand{\Yes}{\cmark}
\newcommand{\No}{\phantom{\cmark}}
\definecolor{Block}{RGB}{245,247,250}
\begin{table}[t]
    \centering
    \small 
    \caption{Commonly used strategies in terms of their used design patterns and tools. CI stands for code interpreter. Pipelines do not use conditionals, workflows are any strategy using conditionals, and agents are strategies using recursion.}
    \label{tab:strategies}
    \begin{tabular}{lcccc}
    \toprule
         Name & Parallelization & Conditionals & Recursion & Tools (base processes)\\
         \midrule
         \rowcolor{Block}
         \textbf{Pipelines} & -- & \xmark & -- & -- \\
         CoT \citep{wei2022chain} & \No & \No & \No & LLM\\
         Self-Consistency \citep{wang2023selfconsistency} & \Yes & \No & \No & LLM\\
         \rowcolor{Block}
         \textbf{Workflows} & -- & \textbf{\Yes} & -- & -- \\
         Code \citep{chen2023program} & \No & \Yes & \No & LLM, CI\\
         \rowcolor{Block}
         \textbf{Agents} & -- & -- & \Yes & -- \\
         Eval-Opt \citep{madaan2023self} & \No & \Yes & \Yes & LLM\\
         CodeAct \citep{codeact} & \No & \Yes & \Yes & LLM, CI\\
         \bottomrule
    \end{tabular}
\end{table}

Figure~\ref{fig:strategy-diffs} shows that different strategies result in significantly different behavior in terms of cost and accuracy. Among strategies we consider, the Code strategy is best for 3-SAT and Word Sorting, but worst for AIME and Movie Recommendation. Even among strategies with comparable accuracy, cost can vary dramatically: Eval-Opt generally achieves similar accuracy to Self-Consistency but often at half the cost. Notably, more expressive strategies do not guarantee better performance: while agentic strategies can theoretically emulate simpler strategies through their conditional and recursive structure, they may fail to select the right behavior in practice and, even when successful, often incur substantially higher computational costs than using the simpler strategy directly (for example, CodeAct is not the most effective strategy in Figure~\ref{fig:strategy-diffs}, despite being the most general). These observations highlight a fundamental trade-off where practitioners must balance solution quality against computational expense, often termed ``overthinking.''

\subsection{Incorporating Feedback to Strategies}

The observations above motivate systems that can adapt their strategy selection to each problem. Ideally, such systems should improve with experience: using information from past problem-solving attempts to make better strategy choices on future problems. This requires maintaining a state during inference that captures which strategies work well for which types of problems.

Existing work addresses two separate problems but not both simultaneously. Some work focuses on maintaining state during testing but is limited in what the state can influence. These methods add state to strategies compiled before inference, so they are limited to steering strategies by changing their input \citep{dynamic-cheat, mem0}. Other work adapts general strategies to experience but requires expensive offline optimization to compile an optimized strategy that is then left stateless during inference \citep{adas, cheng2024trace}. Figure~\ref{fig:overview} illustrates how training-based optimization methods (b) compare to state-based adaptation approaches (a) and our approach (c).

The success of offline strategy optimization in modifying entire strategies motivates a method that compiles useful strategies on the fly by learning from experience. This is the approach we take and introduce in the following section.
\section{Experience-Guided Reasoner (\ourabbrev)}
\label{sec:method}

In this section, we introduce a system that generates tailored strategies dynamically at inference time by learning from accumulated experience. Our approach centers on two key components: a \textit{\guide} that generates complete, problem-specific strategies based on past experience, and a \textit{\updater} that processes execution outcomes to improve future strategy generation.

\subsection{General Design}

The \textbf{\guide} generates complete strategy specifications tailored to each problem based on the current context and past experiences. Rather than making decisions step-by-step like traditional agents, it produces a complete computational procedure upfront--specifying LLM calls, sampling parameters, tools, and control flow. The \textbf{\updater} processes execution outcomes, including reasoning traces and verifier feedback, maintaining a structured memory that improves strategy generation.

\newcommand{\mycomment}[1]{\hspace{0.5cm} \textcolor{gray}{\small // #1}}
\begin{algorithm}[t]
\caption{\ourmethod}
\label{alg:method}
\begin{algorithmic}[1]
\REQUIRE Question $q$, context $\Sigma$, and exploration factor $k$.
\ENSURE Answer $a$ and updated context $\Sigma'$.
\STATE $\{S_1, \dots, S_k\} \gets \textbf{\guide}(q, \Sigma, k)$ \mycomment{generate $k$ candidate strategies}
\STATE $e = []$ \mycomment{initialize experience collection}
\FOR{$i = 1$ \TO $k$ in \textbf{parallel}}
    \STATE $\sigma \gets \text{null}$ \mycomment{initialize strategy state}
    \STATE $(a_i, t_i, c_i) \gets \textbf{Execute}(S_i, q, \sigma)$ \mycomment{execute strategy $S_i$}
    \STATE $f_i \gets \text{verifier}(q, a_i)$ \mycomment{get verifier feedback}
    \STATE $e \gets e + [(q, a_i, t_i, c_i, f_i)]$ \mycomment{collect experience tuple}
\ENDFOR
\STATE $\Sigma' \gets \textbf{\updater}(e, \Sigma)$ \mycomment{update context with experiences}
\STATE $a \gets a_1$ \mycomment{use answer from first strategy}
\RETURN $a, \Sigma'$
\end{algorithmic}
\end{algorithm}
Formally, these components have types:
\begin{align*}
    \text{\guide}: \proc{\Sigma}{\text{Str}}{\left(\proc{\sigma}{\text{Str}}{\text{Str}}\right)} \qquad \text{\updater}: \proc{\Sigma}{\text{List[Exp]}}{\Sigma}
\end{align*}
where Exp represents (question, answer, trace, cost, feedback) tuples. The state space is divided into $\Sigma$ (between-episode memory maintained by \updater) and $\sigma$ (within-episode memory used during strategy execution). The \guide has a state $\Sigma$, takes a string as input and outputs a strategy as defined in \cref{sec:representation} (which itself uses a state $\sigma$ and takes a string as input and outputs a string). The \updater updates $\Sigma$ based on execution experiences.

The complete \ourabbrev process is shown in \cref{alg:method}, and we also provide an equivalent definition using our strategy language in \cref{app:egur-strategy}.

\subsection{Strategy Selection with the \guide}

The \guide generates $k$ candidate strategies for each problem by conditioning on the query $q$ and relevant experiences from the context $\Sigma$. This design provides some key benefits:  \textbf{observability} of the system's behavior with complete strategy specifications; \textbf{exploration} of the space of possible strategies by generating multiple candidates; and \textbf{comparison} of the relative performance of different approaches for future strategy generation.
The exploration factor $k$ balances computational cost with strategy diversity---setting $k=1$ minimizes overhead, while larger values enable discovery of better strategies at increased cost.

We invoke the \guide $k$ times on problem $q$ to obtain $k$ candidate strategies $\{S_1, \ldots, S_k\}$, which we execute to obtain performance measures. Similar to Group Relative Policy Optimization (GRPO) \cite{grpo}, we use relative comparisons within each group of $k$ strategies: the \updater records which strategies perform better relative to alternatives on similar problems, enabling the \guide to improve over time. In practice, the \guide is implemented as an LLM call that produces strategy code, which is then parsed from the response and executed (see \cref{app:consultant} for the prompt).

\subsection{Feedback Integration with the \updater}

The \updater maintains structured memory to guide future strategy generation. Rather than accumulating raw experiences, it performs selective abstraction to maximize information utility while preventing memory bloat. The \updater processes experiences $e = [(q_1, a_1, t_1, c_1, f_1), \ldots, (q_k, a_k, t_k, c_k, f_k)]$ from executed strategies, where each tuple contains the query, answer, execution trace, computational cost, and verifier feedback. 

The context $\Sigma$ maintains two key components: a \textbf{Strategy Library} that stores successful strategies with their problem characteristics for reuse and template generation, and \textbf{General Notes} that capture high-level insights about strategy effectiveness, common failure patterns, useful techniques, and problem-solving heuristics. The entire context is passed directly to the \guide for each problem, 
enabling it to perform retrieval and synthesis in-context rather than requiring external retrieval mechanisms.

To prevent unbounded memory growth, the \updater implements selective retention policies, prioritizing recent experiences and frequently accessed patterns while removing outdated information to maintain responsiveness to evolving problem distributions while preserving valuable learned knowledge. In practice, the \updater is implemented as an LLM call that produces memory insertions and deletions which are parsed and applied (see \cref{app:contextupdater} for the prompt).

\section{Experiments}

We evaluate \ourabbrev across diverse reasoning tasks to answer four key research questions: (RQ1) Does it outperform baselines in accuracy and cost? (RQ2) Does dynamic strategy generation improve over existing stateful methods? (RQ3) Is comparative strategy evaluation important for continual improvements? (RQ4) Does \ourabbrev learn novel and useful strategies from experience?

\subsection{Experimental Setup}

\paragraph{Datasets.} We evaluate on five datasets: AIME (2022-2024 for training, 2025 held-out), 3-SAT (5-40 variables, 400 training/40 test samples), and three BBEH \citep{bbeh} tasks (movie recommendation, word sorting, object counting, 100 training/100 test each). We process training data in shuffled batches of ten samples, evaluating all ten in parallel then sequentially updating based on each sample's experience.

\paragraph{Models.} Claude 3.7 Sonnet, Qwen3-Next-80B-A3B-Thinking, and GPT-OSS-120B.

\paragraph{Baselines.} CodeAct \citep{codeact} (stateless code-interpreter agent), Dynamic Cheatsheet \citep{dynamic-cheat} (CodeAct with string-based memory appended to inputs), and Mem0 (CodeAct with vector database memory). We modify Dynamic Cheatsheet to incorporate verifier feedback during updates.

\paragraph{Verifiers.} For all tasks except for 3-SAT, we use ground-truth based verifiers that compare the final answer against the correct solution. For 3-SAT, we verify the proposed assignment satisfies all clauses using a satisfiability checker. The verifier provides only binary feedback (correct/incorrect) which is passed to the \updater.

\paragraph{Implementation details.} All LLM calls use temperature 0 and a maximum token budget of 20,000 tokens. For Claude 3.7 Sonnet, we enable extended thinking mode with a maximum thinking budget of 10,000 tokens; when thinking mode is enabled, Claude's temperature is automatically set to 1.0. These settings apply to all strategy executions, \guide calls, and \updater calls across all methods. With \ourabbrev, the strategy can include LLM calls without thinking mode or with non-zero temperature.

\paragraph{Evaluation.} We report prequential accuracy (accuracy before updating on each sample) and cumulative cost for strategy execution (excluding system feedback and update costs). Continual updates use one seed; held-out evaluation uses three seeds. Costs are in USD based on API pricing.

\subsection{RQ1: Does \ourabbrev outperform baselines in accuracy and cost?}

\begin{figure}
    \centering
    \includegraphics[width=\linewidth]{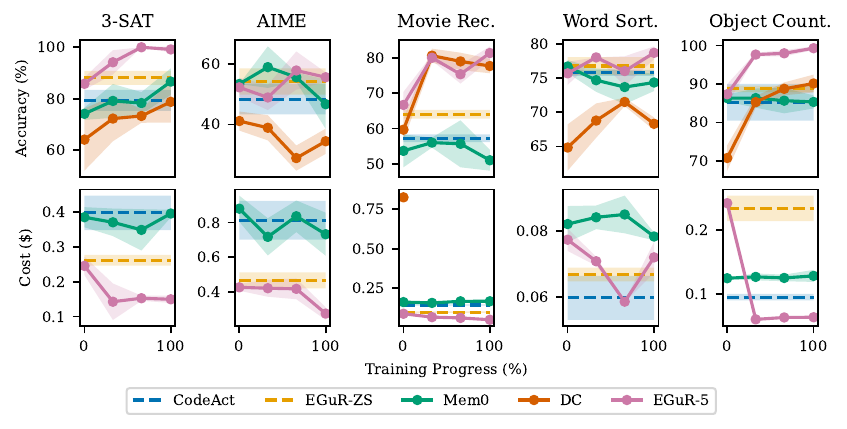}
    \caption{Evolution of accuracy and cost on held-out evaluation sets as training progresses for Claude 3.7 Sonnet. \ourabbrev-5 consistently improves accuracy while reducing cost with more experience. Cost is shown up to \$1.0 for visualization; Dynamic Cheatsheet (DC) typically exceeds this threshold, reaching \$9.95, \$2.26, \$2.88, \$4.32, and \$7.16 per sample after training on 3-SAT, AIME, Movie Rec., Word Sort., and Object Count., respectively. The complete results for Claude 3.7 Sonnet, GPT-OSS-120B, and Qwen3-Next-80B-A3B-Thinking are included in \cref{app:full-results}.}
    \label{fig:holdout}
\end{figure}

\Cref{fig:holdout} shows \ourabbrev's performance on held-out evaluation sets as training progresses. Two key trends emerge: accuracy increases with more experience, and cost decreases as \ourabbrev learns more efficient strategies. Notably, Dynamic Cheatsheet's costs are significantly higher than other approaches (often exceeding \$1.00 per sample) due to the increasing context length from accumulating memory, and the underlying strategy taking many turns before outputting a final answer. In contrast, \ourabbrev maintains low and decreasing costs through selective memory management. Mem0 shows minimal cost reduction and modest accuracy improvements, suggesting limited adaptation capability.

\cref{tab:prequential-eval} provides detailed prequential accuracy and cost results across all datasets and models. \ourabbrev consistently achieves the best accuracy-cost trade-offs: on Claude 3.7 Sonnet, \ourabbrev-5 achieves 96.0\% accuracy on 3-SAT at \$0.152 cost versus CodeAct's 77.0\% at \$0.257 and Dynamic Cheatsheet's 89.9\% at \$76.353. Similar patterns hold across models and tasks.

\subsection{RQ2: Does dynamic strategy generation improve over existing stateful strategies?}

The key distinction between \ourabbrev and existing stateful methods is architectural: \textbf{\ourabbrev generates a complete strategy specification for each problem before execution}, while existing methods \textbf{steer dynamic strategies at runtime by modifying their inputs}. Concretely, Mem0 and Dynamic Cheatsheet prepend memory to a fixed CodeAct agent's inputs, influencing its behavior indirectly. In contrast, \ourabbrev produces entirely new strategy specifications---including prompts, sampling parameters (temperature, max tokens), tool availability, and control flow---tailored to each problem.

This architectural difference has profound implications. Methods that steer via input modification cannot remove tools, change sampling parameters, or switch between agentic and workflow paradigms. \ourabbrev is significantly more flexible: for instance, it learns to completely remove the code interpreter tool when it harms performance, adjust temperature from 0.7 to 0.0 for deterministic tasks, and replace multi-turn agentic strategies with single-call workflows when appropriate.

The results in \cref{fig:holdout} reflect these differences. Mem0 achieves minimal cost reduction despite maintaining memory, as it cannot fundamentally change the underlying strategy's computational structure. Dynamic Cheatsheet's costs grow unboundedly as accumulated memory increases context length in every call. \ourabbrev achieves superior accuracy while reducing costs by 2-111× through complete strategy adaptation.

\subsection{RQ3: Is comparative strategy evaluation important for continual improvements?}

\begin{figure}[t]
    \centering
    \includegraphics[width=\linewidth]{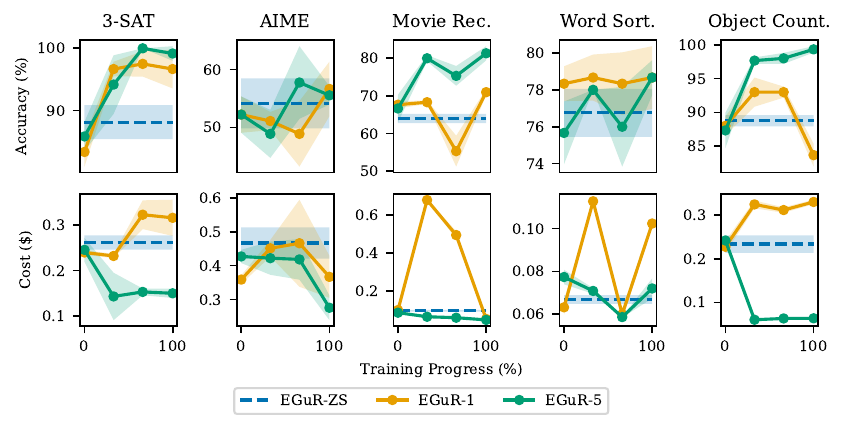}
    \caption{Ablation of exploration level in \ourabbrev for Claude 3.7 Sonnet. Higher exploration levels (more strategies per problem) generally improve both accuracy and cost-efficiency by enabling comparative evaluation of strategy effectiveness.}
    \label{fig:claude-ablation}
\end{figure}

We ablate the exploration level $k$ (strategies generated per problem) to assess the importance of comparative evaluation. \Cref{fig:claude-ablation} shows results for Claude 3.7 Sonnet. \ourabbrev-ZS (zero-shot, no memory updates) serves as a stateless baseline, \ourabbrev-1 generates one strategy per problem and learns from absolute feedback, while \ourabbrev-5 generates five strategies enabling comparative evaluation.

Three findings emerge. First, memory is essential: \ourabbrev-1 consistently outperforms \ourabbrev-ZS, showing that learning from experience helps even without comparison. Second, comparative evaluation provides substantial additional benefits: \ourabbrev-5 outperforms \ourabbrev-1 on most tasks, with large improvements on 3-SAT and object counting. Third, cost decreases with higher $k$ because comparative evaluation helps discover efficient strategies faster. 

\subsection{RQ4: Does \ourabbrev learn novel and useful strategies from experience?}

We analyze strategies generated before and after training to identify learned adaptations. \ourabbrev exhibits several consistent patterns: for CodeAct strategies, it learns to specify allowed libraries, include useful code snippets, and add error handling; more generally, it increases specificity when general approaches fail but simplifies when appropriate.

We also find that the code interpreter tool sometimes harms performance while increasing cost. For BBEH object counting, intuition suggests CodeAct would excel (problems involve adding many numbers), but \ourabbrev converges on a single LLM call with detailed instructions which proves more accurate and substantially cheaper. The learned strategy includes specific guidance for parsing text, categorizing items, and handling quantity changes. The strategy is shown in \cref{app:strat-qual}.

Similarly, for BBEH word sorting, \ourabbrev learns to distinguish algorithmic sorting problems (which benefit from Python's \texttt{sort}) from reasoning problems (identifying logical mistakes in explanations). For the latter, it adopts CoT with verifier-based fallback rather than CodeAct.

These findings show \ourabbrev learns meaningful heuristics for when to use tools versus LLM-based reasoning, when to increase versus decrease computational investment, and how to tailor strategies through prompts to problem characteristics. Additional examples appear in \cref{app:strat-qual}.
\section{Related Work}

\paragraph{Prompting strategies.}  
Prompting methods including Chain-of-Thought (CoT) \citep{wei2022chain}, Self-Consistency \citep{wang2023selfconsistency}, Program-of-Thoughts (PoT) \citep{chen2023program}, and many other approaches \citep{lyu2023faithful, react, leasttomost, tot} elicit intermediate reasoning traces but rely on fixed, non-adaptive scaffolds. More recent approaches like CodeAct \citep{codeact}, Self-Discover \citep{dynamic-cheat}, and Meta-Prompting \citep{meta-prompting} dynamically compose solution solving methods per instance, but still do not leverage past experience or adapt their strategies over time. These methods are thus dynamic at inference but remain stateless across instances.

\paragraph{Prompt adaptation from inference-time state or training.}  
Another line of work adapts prompts or parameters within fixed agent workflows. DSPy \citep{dspy} and TextGrad \citep{textgrad} are systems to optimize prompts or even the inputs to fixed strategy structures using a training dataset with labels or an answer verifier, while Dynamic Cheatsheet \citep{dynamic-cheat} and Buffer of Thoughts \citep{bot} accumulate and retrieve artifacts during inference for instance-level adaptation.
Concurrent to this work, training-free GRPO \citep{cai2025training} was proposed as an effective method for online prompt optimization which is similar to our GRPO inspired approach to producing effective memory updates from the \updater, and Agentic Context Engineering \citep{zhang2025agentic} improves over Dynamic Cheatsheet by providing more structure for memory updates.
Systems such as DSPy are highly general and allow for many different algorithms for offline prompt optimization \citep{gepa, opro, ape}.
Other methods such as Per-Instance Program Synthesis (PIPS) \citep{pips}, model routing \citep{frugalgpt, mop, routellm}, and proprietary products like GPT-5 \citep{gpt5} rely on trained routers to select between a fixed set of strategies on a per-instance basis.
These approaches achieve stronger performance through experience but are limited in their adaptability based on only modifying the input to a fixed strategy or selecting between a fixed number of strategies.

\paragraph{Full strategy adaptation.}  
Beyond prompt tuning, ADAS \citep{adas} treats agent design as a search problem over Python programs defining agents, enabling strategy-level adaptation. However, ADAS and similar approaches \citep{zhang2025aflow, zhuge2024gptswarm, li2024autoflow, zhang2025evoflow} operate offline and at the task level, requiring multiple episodes to converge. In contrast, our method introduces an \emph{online adaptive meta-agent} that performs dynamic, instance-level adaptation and test-time strategy search, combining the benefits of prior work on dynamic strategies, memory-based learners, and automated design in a unified framework.
\section{Conclusion}

We have described a system, \ourabbrev, which generates complete strategies---computational procedures involving LLM calls, tools, and control logic---dynamically at inference time based on accumulated experience. Specifically, a \guide component generates problem-specific strategy specifications conditioned on past experience, while a \updater maintains structured memory of strategy effectiveness. Unlike existing adaptive systems that use memory for textual steering of an LLM or agent, our approach has much greater flexibility, enabling adaptation of all strategy components (prompts, sampling parameters, tool configurations, and control logic). Across five benchmarks, \ourabbrev achieves up to 14\% relative accuracy improvements while reducing computational costs by up to 111x, with both metrics improving as the system gains experience. Our analysis reveals \ourabbrev learns meaningful heuristics: when to deploy expensive agentic strategies versus lightweight workflows, when tools help versus harm, and how to tailor computational investment to problem difficulty. 

Our approach has some limitations that suggest directions for future work. First, \ourabbrev uses ground truth feedback from verifiers to learn effectively. Exploring whether this feedback can be replaced with weaker signals---such as LLM-based evaluation---is an important direction. Second, the effectiveness of the system hinges on the \guide's zero-shot strategy generation capabilities, which may be suboptimal for unfamiliar problem types. In this case, it may be necessary to train or optimize the Guide through reinforcement learning or other training methods. Similarly, the \updater relies on an LLM to manage memory, which may not optimally balance memory size against information utility, and may benefit from meta-learning approaches for more effective memory management.

\bibliography{refs}
\bibliographystyle{unsrtnat}

\appendix

\section{Related Work Taxonomy}
A taxonomy of the related work is shown in \cref{tab:taxonomy}.

\definecolor{Block}{RGB}{245,247,250}

\begin{table}[t]
\centering
\begingroup
\setlength{\tabcolsep}{4pt}        %
\renewcommand{\arraystretch}{1.05} %
\caption{Taxonomy of related work by \textbf{S@T}=Stateful at test, \textbf{Adapted}=What is adapted in response to the state, input, or training samples, \textbf{Training-free}=If the method does not require an initial training stage}
\label{tab:taxonomy}
\scriptsize

\begin{tabular}{lccc}
\toprule
\textbf{Method} & \textbf{S@T} & \textbf{Adapted} & \textbf{Training-free}\\
\midrule
\rowcolor{Block}
\multicolumn{4}{l}{\textbf{Stateless strategies (human designed)}} \\
CoT                                 & \No & & \Yes\\
Self-Consistency                    & \No & & \Yes\\
PoT (non-iterative)                 & \No & & \Yes\\
CodeAct \citep{codeact}             & \No & & \Yes\\
\midrule
\rowcolor{Block}
\multicolumn{4}{l}{\textbf{Adapts prompts from state/training}} \\
Self-Discover \citep{dynamic-cheat}      & \No & Prompt & \Yes  \\
DSPy \citep{dspy}                        & \No & Prompts  & \No \\
TextGrad \citep{textgrad}                & \No & Prompts, Inputs  & \No \\
Buffer of Thoughts \citep{bot}           & \Yes & Prompt & \Yes \\
Dynamic Cheatsheet \citep{dynamic-cheat} & \Yes & Prompt & \Yes \\
\midrule
\rowcolor{Block}
\multicolumn{4}{l}{\textbf{Full strategy adaptation}} \\
FlowReasoner \citep{flowreasoner}     & \No  & Strategy & \No  \\
ADAS \citep{adas}                     & \No  & Strategy  & \No \\
\ourabbrev                            & \Yes & Strategy & \Yes \\
\bottomrule
\end{tabular}
\endgroup
\end{table}

\section{Strategies}
\label{app:strategies}
\subsection{Syntax}
The syntax of strategies is separated into values, expressions and programs.

\paragraph{Values}
\[
\begin{array}{rcl}
v & \bnfdef & \text{str} \bnfalt \text{float} \bnfalt \text{true} \bnfalt \text{false} \bnfalt \text{null}\\
&\bnfalt& [v\;\; (,\; v)*]\\
&\bnfalt& \{\text{str}: v\;\; (,\; \text{str}:v)*\}\\
\end{array}
\]
Values are the typical Python values including strings, floats, booleans, the None value (written null) as well as lists and dictionaries.
 
\paragraph{Expressions}
\[
\begin{array}{rcl}
e & \bnfdef & x \bnfalt v \bnfalt e[e] \bnfalt e \doubleplus e \bnfalt e \oplus e \bnfalt e + e \bnfalt e - e \bnfalt e * e \bnfalt e \div e\\
&\bnfalt& \kw{lambda}\;x.\; e \bnfalt e(e)
\end{array}
\]
Expressions consist of either a variable $x$, a value, or various operations of expressions including list indexing, list append $\doubleplus$, dictionary append $\oplus$, float arithmethic $(+, -, *, \div)$, lambda functions, and function application.

\paragraph{Programs}
\[
\begin{array}{rcl}
p & \bnfdef & \text{base\_proc} \bnfalt \kw{return} \bnfalt \kw{pure}\;e \bnfalt \kw{get} \bnfalt \kw{put}\;e\\
& \bnfalt & p; p\\
& \bnfalt & p \parallel p\\
& \bnfalt & \kw{if}\;p\;\kw{then}\;p\;\kw{else}\;p\\
& \bnfalt & \kw{fix}\;p.\;p
\end{array}
\]
Programs are then either a base process, a return statement (which directly returns the input), an expression lifted to a program using $\kw{pure}$, a $\kw{get}$ which returns the current state of the program, a $\kw{put}$ of an expression which updates the state with the expression, or a sequential or parallel composition of programs, a conditional of programs, or a fixpoint of a program.

\subsection{Semantics}
\label{app:semantics}

We now provide a denotational semantics for the strategy language. First, a program in this language is a \textit{stateful process} mapping between two sets:
\begin{align*}
    \text{Proc}(A, B, \Sigma): (A\times \Sigma)\to (B\times \Sigma)
\end{align*}
where $\Sigma$ is the state space of the process. Intuitively, a process can be thought of as a mapping from $A$ to $B$ which has access to some state $\Sigma$. To simplify how we compose processes, we use the state monad:
\begin{align*}
    \text{State}(A): \Sigma\to (A\times \Sigma)
\end{align*}
and then rewrite the process type as $\text{Proc}(A, B, \Sigma): A\to \text{State}(B)$ which can be thought of as a function which maps an input in $A$ to a computation which when given a state will produce an output in $B$ and an updated state. The choice of the state space $\Sigma$ determines how processes share information when composed as well as allows for tracking information such as execution traces, cost, and strategy structure. For instance, we use $\Sigma = M\times P\times\text{Trace}\times \text{Cost}$ where $M$ is the current conversation log of any model calls, $P$ is the execution state for any code execution, Trace is the current execution trace, and Cost is the current execution cost of the strategy.

The program meaning function $\den{\cdot}$ is now a mapping from the syntactic forms defined above to processes.

\[
\llbracket \cdot \rrbracket \;:\; \mathsf{Program} \;\longrightarrow\; \text{Proc}(A, B, \Sigma)
\]
\[
\begin{array}{rcl}
\den{\text{base\_proc}} &=& \text{base\_proc}, \\[0.35em]
\den{\kw{return}} &=& \lambda\ a.\; \lambda c.\; (a, c), \\[0.35em]
\den{\kw{pure}\;f} &=& \lambda\ a.\; \lambda c.\; (\den{f}(a), c), \\[0.35em]
\den{\kw{get}} &=& \lambda\ a.\; \lambda c.\; (c, c), \\[0.35em]
\den{\kw{put}\;e} &=& \lambda\ a.\; \lambda c.\; (a, \den{e}), \\[0.35em]
\den{p_1 ; p_2} &=& \lambda\ a.\; \lambda c.\; \text{let}\;(b, c_1)\leftarrow \den{p_1}(a)(c)\\
&&\text{in}\;\den{p_2}(b)(c_1), \\[0.35em]
\den{p_1 \parallel p_2}(a)(c) &=& \lambda\ a.\; \lambda c.\; \text{let}\;(b_1, c_1)\leftarrow \den{p_1}(a)(c)\;\text{and}\;(b_2, c_2)\leftarrow \den{p_2}(a)(c),\\
&&\text{in}\;((b_1, b_2), (c_1, c_2)),\\[0.35em]
\llbracket \kw{if}\;p_1\;\kw{then}\;p_2\;\kw{else}\;p_3 \rrbracket(s) &=&
    \lambda\ a.\; \lambda c.\; \text{let}\;(b, c_1)\leftarrow \den{p_1}(a)(c)\\
    &&\text{in}
    \begin{cases}
    \den{p_2}(a, c_1), & \text{if } b = \mathsf{true} \\
    \den{p_3}(a, c_1), & \text{if } b = \mathsf{false}
    \end{cases},\\[1em]
\llbracket \mathbf{fix} \, p_1 .\, p_2 \rrbracket &=& \operatorname*{lfp}_{\sqsubseteq} \;
    \bigl( p \mapsto \llbracket p_2 \rrbracket_{p_1 \mapsto p} \bigr).
\end{array}
\]

\subsection{\guide and \updater}

The \guide is a strategy which takes a query as input and outputs a strategy for solving the problem, as well as has access to a state which it can use to store information between queries. Therefore, the \guide is just a higher-order strategy, or a stateful mapping from inputs to strategies and its type is the following:
\begin{align*}
    \text{\guide}(A, B, \Sigma): \text{Proc}(A, \text{Proc}(A, B, \sigma), \Sigma).
\end{align*}

The memory updater can also be written as a strategy which takes the current experience, the current running state, and outputs the updated state:
\begin{align*}
    \text{\updater}(\text{Exp}, \Sigma): \text{Proc}(\text{Exp}, \Sigma, \Sigma).
\end{align*}

\subsection{Strategy Cost}
\label{app:cost-def}
We define the cost of a strategy recursively based on the costs of the base processes $P\in\mathcal{P}$ defined by $c_P(s)\in\mathbb{R}_{\ge 0}$, the cost of running $P$ on input $s$.

\begin{align*}
\mathrm{Cost} &: \mathsf{Term} \times S \longrightarrow \mathbb{R}_{\ge 0} \cup \{\infty\}, \\[0.5em]
\mathrm{Cost}(t,s) &=
\begin{cases}
c_P(s), 
& \text{if } t = P \in \mathcal{P}, \\[0.5em]
\mathrm{Cost}(t_1, s) +
\displaystyle \mathbb{E}_{s' \sim \llbracket t_1 \rrbracket(s)}
      \bigl[ \mathrm{Cost}(t_2, s') \bigr],
& \text{if } t = t_1 ; t_2, \\[1em]
\mathrm{Cost}(t_1, s) + \mathrm{Cost}(t_2, s),
& \text{if } t = t_1 \,\Vert\, t_2, \\[0.5em]
\begin{cases}
\mathrm{Cost}(t_1, s), & b(s) = \mathsf{true}, \\
\mathrm{Cost}(t_2, s), & b(s) = \mathsf{false},
\end{cases}
& \text{if } t = \text{if } b \text{ then } t_1 \text{ else } t_2, \\[1em]
\mathrm{Cost}\bigl(t[X \mapsto \mathbf{fix}\,X.\,t], s\bigr),
& \text{if } t = \mathbf{fix}\,X.\,t \quad\text{(least fixed point)}.
\end{cases}
\end{align*}

\subsection{\ourabbrev Implemented as a Strategy}
\label{app:egur-strategy}
We show how \ourabbrev is represented as a strategy in \cref{alg:method-strategy} where we make sequential and parallel composition of processes expliclit.

\begin{algorithm}[t]
\caption{\ourmethod}
\label{alg:method-strategy}
\begin{algorithmic}[1]
\REQUIRE Question $q$, context $\Sigma$, and exploration factor $k$.
\ENSURE Answer $a$ and updated context $\Sigma'$.
\STATE \textbf{update} (\textbf{fun} $(q, \Sigma)$: $\Sigma + \{\text{``question''}: q\}$); \qquad \mycomment{store the question in context}
\STATE $\textbf{\color{blue}\guide}_k$; \qquad \mycomment{generate $k$ strategies conditioned on the question and context}
\STATE (\textbf{recfun} ExecuteStrategies: \qquad \mycomment{execute all strategies in parallel}
\STATE \qquad \textbf{if} \textit{IsEmpty}
\STATE \qquad \textbf{then} \textbf{return}
\STATE \qquad \textbf{else} (\textit{GetFirst}; {\color{blue}\textbf{Trace}}; \textit{Verify}) $\parallel$ (\textit{PopFirst}; ExecuteStrategies)
\STATE );
\STATE \textbf{update} (\textbf{fun} $(e, \Sigma)$: $\Sigma + \{\text{``answer''}: e[0][0]\}$); \mycomment{store the answer from the first strategy}
\STATE \textbf{\color{blue}\updater} \qquad \mycomment{update the context from the $k$ experiences}
\STATE \textbf{get} ``answer''
\end{algorithmic}
\end{algorithm}

\subsection{Common Strategies}
\label{app:common-strats}

In this section, we describe the five strategies shown in \cref{tab:strategies}. We also show the implementations of the strategies in our strategy language in \cref{fig:additional-strats}.

The CoT strategy is a single LLM inference call without any tools, the Code strategy is a single round of code writing and executing the code with an interpreter to produce the answer, the Eval-Opt strategy is an evaluator-optimizer loop where the optimizer produces candidate solutions and the evaluator critiques the solutions, the Self-Consistency strategy samples 10 solutions from the CoT strategy and then returns the majority answer, and the CodeAct strategy uses multiple rounds of code generation and execution to solve the problem, and is prototypical of an \textit{agentic} workflow.

\begin{figure*}[t]
\centering

\begin{subfigure}[b]{0.16\textwidth}
\centering
\begin{tabular}{@{}r@{\hspace{0.5em}}l@{}}
\textcolor{gray}{\footnotesize 1} & $\text{CallLLM}$ \\[0.2em]
\end{tabular}
\caption{CoT}
\end{subfigure}
\hfill
\begin{subfigure}[b]{0.26\textwidth}
\centering
\begin{tabular}{@{}r@{\hspace{0.5em}}l@{}}
\textcolor{gray}{\footnotesize 1} & $(\text{CallLLM} \parallel \text{CallLLM} \parallel$ \\[0.2em]
\textcolor{gray}{\footnotesize 2} & $\phantom{(}\text{CallLLM} \parallel \text{CallLLM} \parallel$ \\[0.2em]
\textcolor{gray}{\footnotesize 3} & $\phantom{(}\text{CallLLM});$ \\[0.2em]
\textcolor{gray}{\footnotesize 4} & $\text{MajorityVote}$ \\[0.2em]
\end{tabular}
\caption{Self-Consistency}
\end{subfigure}
\hfill
\begin{subfigure}[b]{0.30\textwidth}
\centering
\begin{tabular}{@{}r@{\hspace{0.5em}}l@{}}
\textcolor{gray}{\footnotesize 1} & $\text{CallLLM};$ \\[0.2em]
\textcolor{gray}{\footnotesize 2} & $(\mathbf{if}\ \text{ContainsCode}$ \\[0.2em]
\textcolor{gray}{\footnotesize 3} & $\phantom{(}\mathbf{then}\ \text{ExecCode}$ \\[0.2em]
\textcolor{gray}{\footnotesize 4} & $\phantom{(}\mathbf{else}\ \mathbf{return})$ \\[0.2em]
\end{tabular}
\caption{Code}
\end{subfigure}
\hfill
\begin{subfigure}[b]{0.24\textwidth}
\centering
\begin{tabular}{@{}r@{\hspace{0.5em}}l@{}}
\textcolor{gray}{\footnotesize 1} & $\mathbf{recfun}\ \text{Eval-Opt}:$ \\[0.2em]
\textcolor{gray}{\footnotesize 2} & $\quad \text{CallOptLLM};\;$ \\[0.2em]
\textcolor{gray}{\footnotesize 3} & $\quad (\mathbf{if}\ \text{EvalLLM}$ \\[0.2em]
\textcolor{gray}{\footnotesize 4} & $\phantom{(}\quad \mathbf{then}\ \mathbf{return}$ \\[0.2em]
\textcolor{gray}{\footnotesize 5} & $\phantom{(}\quad \mathbf{else}\ \text{Eval-Opt})$
\end{tabular}
\caption{Eval-Opt}
\end{subfigure}

\caption{Implementation of CoT, Self-Consistency, Code, and Eval-Opt as strategies}
\label{fig:additional-strats}
\end{figure*}

\section{Full results}
\label{app:full-results}

The prequential evaluation results for all models is included in \cref{tab:prequential-eval}.

\begin{table}[ht]
\centering
\footnotesize
\caption{Prequential Accuracy (\%) and Cost (\$) on each dataset. \ourabbrev consistently achieves the best accuracy-cost trade-offs, with bold indicating best accuracy and underline indicating second-best. Ours-$k$ denotes \ourabbrev with exploration level $k$.}
\label{tab:prequential-eval}
\begin{tabular}{l l r r r r r r r r r r}
\toprule
\multirow{2}{*}{\textbf{Model}} & \multirow{2}{*}{\textbf{Method}} & \multicolumn{2}{c}{\textbf{3-SAT}} & \multicolumn{2}{c}{\textbf{AIME}} & \multicolumn{2}{c}{\textbf{Movie Rec.}} & \multicolumn{2}{c}{\textbf{Word Sort}} & \multicolumn{2}{c}{\textbf{Object Count}}\\
\cmidrule(lr){3-4} \cmidrule(lr){5-6} \cmidrule(lr){7-8} \cmidrule(lr){9-10} \cmidrule(lr){11-12}
& & \textbf{Acc} & \textbf{Cost} & \textbf{Acc} & \textbf{Cost} & \textbf{Acc} & \textbf{Cost} & \textbf{Acc} & \textbf{Cost} & \textbf{Acc} & \textbf{Cost}\\
\midrule
\multirow{4}{*}{Claude}
& CodeAct & 0.770 & \underline{0.257} & \textbf{0.611} & \underline{0.478} & 0.570 & \underline{0.137} & \underline{0.770} & \textbf{0.067} & 0.800 & \underline{0.098} \\
& CodeAct+Mem0 & 0.860 & 0.325 & \underline{0.589} & 0.548 & 0.540 & 0.141 & 0.760 & 0.085 & \underline{0.870} & 0.174 \\
& DynamicCheatsheet & \underline{0.899} & 76.353 & 0.400 & 8.615 & \textbf{0.840} & 9.152 & 0.700 & 8.231 & 0.870 & 24.895 \\
& \ourabbrev-5 & \textbf{0.960} & \textbf{0.152} & 0.589 & \textbf{0.292} & \underline{0.700} & \textbf{0.067} & \textbf{0.800} & \underline{0.073} & \textbf{0.960} & \textbf{0.075} \\

\midrule
\multirow{4}{*}{Qwen3}
& CodeAct & \textbf{0.875} & \underline{0.032} & 0.622 & \underline{0.010} & \underline{0.680} & \underline{0.002} & \textbf{0.700} & \underline{0.003} & \underline{0.400} & \underline{0.017} \\
& CodeAct+Mem0 & \underline{0.875} & 0.037 & \underline{0.656} & 0.011 & \textbf{0.750} & 0.003 & \underline{0.680} & 0.004 & 0.360 & 0.019 \\
& DynamicCheatsheet & 0.125 & 0.570 & 0.467 & 0.124 & 0.550 & 0.205 & 0.450 & 0.171 & 0.180 & 0.469 \\
& \ourabbrev-3 & 0.695 & \textbf{0.005} & \textbf{0.767} & \textbf{0.006} & 0.650 & \textbf{0.002} & 0.650 & \textbf{0.001} & \textbf{0.540} & \textbf{0.005} \\

\midrule
\multirow{3}{*}{GPT}
& CodeAct & \textbf{0.655} & 0.236 & \underline{0.778} & \textbf{0.073} & \textbf{0.550} & 0.091 & \textbf{0.780} & \textbf{0.087} & \underline{0.690} & \textbf{0.047} \\
& CodeAct+Mem0 & \underline{0.640} & \underline{0.219} & \textbf{0.800} & \underline{0.078} & \underline{0.530} & \textbf{0.061} & \underline{0.780} & \underline{0.129} & \textbf{0.790} & \underline{0.059} \\
& \ourabbrev-3 & 0.650 & \textbf{0.100} & 0.756 & 0.095 & 0.350 & \underline{0.064} & 0.580 & 0.091 & 0.620 & 0.090 \\

\bottomrule
\end{tabular}
\end{table}

Full results for all methods on all three models and five holdout datasets across four evenly spaced checkpoints during training are shown in \cref{tab:acc-1}, \cref{tab:acc-2}, \cref{tab:cost-1}, \cref{tab:cost-2}, and \cref{tab:cost-3}.

\begin{table}[ht]
\centering
\tiny
\caption{Accuracy (\%) across training progress for different methods on 3-SAT and AIME datasets. Values shown as mean $\pm$ standard deviation across seeds. Claude uses EGuR-5, GPT and Qwen use EGuR-3. \textbf{Bold} indicates best performance within each model, \underline{underlined} indicates second best within each model.}
\label{tab:acc-1}
\begin{tabular}{llcccccccc}
\toprule
Model & Method & \multicolumn{4}{c}{3-SAT} & \multicolumn{4}{c}{AIME} \\
 &  & 0\% & 33\% & 66\% & 100\% & 0\% & 33\% & 66\% & 100\% \\
\midrule
\multirow{5}{*}{Claude-3.5-Sonnet} & CodeAct & 79.4$\pm$4.1 & 79.4$\pm$4.1 & 79.4$\pm$4.1 & 79.4$\pm$4.1 & 48.3$\pm$5.0 & 48.3$\pm$5.0 & 48.3$\pm$5.0 & 48.3$\pm$5.0 \\
 & EGuR-ZS & \textbf{88.1$\pm$2.7} & \underline{88.1$\pm$2.7} & \underline{88.1$\pm$2.7} & \underline{88.1$\pm$2.7} & \textbf{54.2$\pm$4.3} & \underline{54.2$\pm$4.3} & 54.2$\pm$4.3 & \underline{54.2$\pm$4.3} \\
 & Mem0 & 74.2$\pm$2.4 & 79.2$\pm$6.6 & 78.3$\pm$4.2 & 86.7$\pm$5.1 & \underline{53.3$\pm$0.0} & \textbf{58.9$\pm$6.8} & \underline{55.6$\pm$1.6} & 46.7$\pm$8.2 \\
 & DC & 64.1$\pm$12.1 & 72.3$\pm$8.8 & 73.3$\pm$2.7 & 78.8$\pm$8.0 & 41.1$\pm$3.1 & 38.9$\pm$4.2 & 28.9$\pm$4.2 & 34.4$\pm$4.2 \\
 & EGuR-5 & \underline{85.8$\pm$2.4} & \textbf{94.2$\pm$4.7} & \textbf{100.0$\pm$0.0} & \textbf{99.2$\pm$1.2} & 52.2$\pm$3.1 & 48.9$\pm$4.2 & \textbf{57.8$\pm$6.3} & \textbf{55.6$\pm$1.6} \\
\midrule
\multirow{5}{*}{GPT-OSS-120B} & CodeAct & \underline{88.8$\pm$5.7} & \textbf{88.8$\pm$5.7} & \underline{88.8$\pm$5.7} & \textbf{88.8$\pm$5.7} & \textbf{79.6$\pm$6.8} & \textbf{79.6$\pm$6.8} & \textbf{79.6$\pm$6.8} & \textbf{79.6$\pm$6.8} \\
 & EGuR-ZS & 83.1$\pm$2.7 & 83.1$\pm$2.7 & 83.1$\pm$2.7 & 83.1$\pm$2.7 & 65.8$\pm$3.6 & 65.8$\pm$3.6 & 65.8$\pm$3.6 & 65.8$\pm$3.6 \\
 & Mem0 & \textbf{91.7$\pm$1.2} & \underline{86.7$\pm$5.9} & \textbf{91.7$\pm$2.4} & \underline{84.2$\pm$2.4} & \underline{76.6$\pm$3.4} & 60.0$\pm$6.7 & 69.1$\pm$9.1 & \underline{79.4$\pm$16.1} \\
 & DC & 10.8$\pm$6.6 & 6.7$\pm$2.4 & 19.2$\pm$4.2 & 13.3$\pm$4.2 & 45.6$\pm$4.2 & 44.4$\pm$5.7 & 53.3$\pm$7.2 & 44.4$\pm$4.2 \\
 & EGuR-3 & 85.8$\pm$2.4 & 80.8$\pm$9.6 & 35.0$\pm$6.1 & 69.2$\pm$12.3 & 65.6$\pm$8.7 & \underline{75.6$\pm$4.2} & \underline{73.3$\pm$7.2} & 76.7$\pm$4.7 \\
\midrule
\multirow{5}{*}{Qwen3-Next-80B} & CodeAct & 63.8$\pm$7.4 & 63.8$\pm$7.4 & 63.8$\pm$7.4 & 63.8$\pm$7.4 & \underline{74.2$\pm$4.9} & 74.2$\pm$4.9 & 74.2$\pm$4.9 & 74.2$\pm$4.9 \\
 & EGuR-ZS & \textbf{80.0$\pm$4.7} & \underline{80.0$\pm$4.7} & \textbf{80.0$\pm$4.7} & \textbf{80.0$\pm$4.7} & \textbf{78.8$\pm$7.5} & \textbf{78.8$\pm$7.5} & \underline{78.8$\pm$7.5} & \textbf{78.8$\pm$7.5} \\
 & Mem0 & \underline{75.0$\pm$0.0} & 52.5$\pm$0.0 & 57.5$\pm$0.0 & \underline{67.5$\pm$0.0} & 66.7$\pm$0.0 & \underline{76.7$\pm$0.0} & \textbf{80.0$\pm$0.0} & \underline{76.7$\pm$0.0} \\
 & EGuR-3 & 72.5$\pm$0.0 & \textbf{90.0$\pm$0.0} & \underline{75.0$\pm$0.0} & 45.0$\pm$0.0 & 56.7$\pm$0.0 & 72.4$\pm$0.0 & 60.0$\pm$0.0 & 55.2$\pm$0.0 \\
\bottomrule
\end{tabular}
\end{table}

\begin{table}[ht]
\centering
\tiny
\caption{Accuracy (\%) across training progress for different methods on Movie Recommendation, Word Sorting, and Object Counting datasets. Values shown as mean $\pm$ standard deviation across seeds. Claude uses EGuR-5, GPT and Qwen use EGuR-3. \textbf{Bold} indicates best performance within each model, \underline{underlined} indicates second best within each model.}
\label{tab:acc-2}
\begin{tabular}{llcccccccccccc}
\toprule
Model & Method & \multicolumn{4}{c}{Movie Rec.} & \multicolumn{4}{c}{Word Sort.} & \multicolumn{4}{c}{Object Count.} \\
 &  & 0\% & 33\% & 66\% & 100\% & 0\% & 33\% & 66\% & 100\% & 0\% & 33\% & 66\% & 100\% \\
\midrule
\multirow{5}{*}{Claude-3.5-Sonnet} & CodeAct & 57.2$\pm$1.1 & 57.2$\pm$1.1 & 57.2$\pm$1.1 & 57.2$\pm$1.1 & 75.8$\pm$0.4 & 75.8$\pm$0.4 & 75.8$\pm$0.4 & 75.8$\pm$0.4 & 85.2$\pm$4.8 & 85.2$\pm$4.8 & 85.2$\pm$4.8 & 85.2$\pm$4.8 \\
 & EGuR-ZS & \underline{64.0$\pm$1.2} & 64.0$\pm$1.2 & 64.0$\pm$1.2 & 64.0$\pm$1.2 & \textbf{76.8$\pm$1.3} & \underline{76.8$\pm$1.3} & \textbf{76.8$\pm$1.3} & \underline{76.8$\pm$1.3} & \textbf{88.8$\pm$0.8} & \underline{88.8$\pm$0.8} & \underline{88.8$\pm$0.8} & 88.8$\pm$0.8 \\
 & Mem0 & 53.7$\pm$4.6 & 56.0$\pm$1.4 & 55.7$\pm$6.6 & 51.0$\pm$2.9 & \underline{76.7$\pm$0.9} & 74.7$\pm$1.7 & 73.7$\pm$2.5 & 74.3$\pm$1.2 & 86.3$\pm$0.9 & 86.3$\pm$2.5 & 85.7$\pm$3.3 & 85.3$\pm$1.7 \\
 & DC & 59.7$\pm$3.1 & \textbf{80.6$\pm$2.0} & \textbf{79.0$\pm$2.5} & \underline{77.7$\pm$2.1} & 64.8$\pm$3.4 & 68.8$\pm$2.5 & 71.5$\pm$0.6 & 68.3$\pm$0.9 & 70.7$\pm$3.3 & 85.3$\pm$1.5 & 88.7$\pm$1.7 & \underline{90.1$\pm$2.3} \\
 & EGuR-5 & \textbf{66.7$\pm$3.7} & \underline{80.0$\pm$0.8} & \underline{75.3$\pm$2.6} & \textbf{81.3$\pm$1.9} & 75.7$\pm$1.7 & \textbf{78.0$\pm$0.0} & \underline{76.0$\pm$2.2} & \textbf{78.7$\pm$0.9} & \underline{87.3$\pm$2.6} & \textbf{97.7$\pm$0.5} & \textbf{98.0$\pm$0.8} & \textbf{99.3$\pm$0.5} \\
\midrule
\multirow{5}{*}{GPT-OSS-120B} & CodeAct & 72.8$\pm$3.3 & \underline{72.8$\pm$3.3} & \underline{72.8$\pm$3.3} & 72.8$\pm$3.3 & \textbf{66.0$\pm$1.7} & \textbf{66.0$\pm$1.7} & \textbf{66.0$\pm$1.7} & \textbf{66.0$\pm$1.7} & \textbf{39.5$\pm$7.1} & 39.5$\pm$7.1 & 39.5$\pm$7.1 & \textbf{39.5$\pm$7.1} \\
 & EGuR-ZS & \textbf{75.0$\pm$3.2} & \textbf{75.0$\pm$3.2} & \textbf{75.0$\pm$3.2} & \textbf{75.0$\pm$3.2} & \underline{64.8$\pm$4.0} & \underline{64.8$\pm$4.0} & 64.8$\pm$4.0 & \underline{64.8$\pm$4.0} & 29.5$\pm$1.8 & 29.5$\pm$1.8 & 29.5$\pm$1.8 & 29.5$\pm$1.8 \\
 & Mem0 & 73.3$\pm$0.9 & 72.0$\pm$2.9 & 72.0$\pm$1.6 & 71.0$\pm$2.4 & 64.0$\pm$3.6 & 63.3$\pm$2.4 & 63.7$\pm$2.1 & 63.7$\pm$1.7 & 35.3$\pm$2.5 & \underline{41.0$\pm$5.4} & \underline{41.0$\pm$2.9} & 38.0$\pm$2.2 \\
 & DC & \underline{73.7$\pm$1.9} & 57.0$\pm$1.4 & 52.3$\pm$0.5 & 59.7$\pm$0.5 & 47.3$\pm$0.5 & 37.7$\pm$4.5 & 45.0$\pm$2.4 & 39.3$\pm$2.5 & 13.0$\pm$4.5 & 15.7$\pm$1.7 & 21.7$\pm$2.1 & 8.7$\pm$5.2 \\
 & EGuR-3 & 62.3$\pm$3.4 & 60.3$\pm$1.7 & 62.3$\pm$3.1 & \underline{74.7$\pm$2.5} & 63.3$\pm$2.6 & 63.0$\pm$2.8 & \underline{65.0$\pm$1.6} & 64.0$\pm$3.6 & \underline{37.3$\pm$2.6} & \textbf{48.3$\pm$4.8} & \textbf{57.3$\pm$1.2} & \underline{39.0$\pm$4.5} \\
\midrule
\multirow{5}{*}{Qwen3-Next-80B} & CodeAct & \underline{46.5$\pm$1.5} & \textbf{46.5$\pm$1.5} & \underline{46.5$\pm$1.5} & \textbf{46.5$\pm$1.5} & \underline{75.5$\pm$2.1} & \textbf{75.5$\pm$2.1} & \underline{75.5$\pm$2.1} & \underline{75.5$\pm$2.1} & \textbf{75.5$\pm$3.2} & \textbf{75.5$\pm$3.2} & \textbf{75.5$\pm$3.2} & \underline{75.5$\pm$3.2} \\
 & EGuR-ZS & 39.2$\pm$1.7 & 39.2$\pm$1.7 & 39.2$\pm$1.7 & 39.2$\pm$1.7 & 64.4$\pm$2.3 & 64.4$\pm$2.3 & 64.4$\pm$2.3 & 64.4$\pm$2.3 & \underline{74.9$\pm$4.5} & \underline{74.9$\pm$4.5} & \underline{74.9$\pm$4.5} & 74.9$\pm$4.5 \\
 & Mem0 & \textbf{47.0$\pm$0.0} & \underline{45.0$\pm$0.0} & \textbf{47.0$\pm$0.0} & \underline{43.0$\pm$0.0} & \textbf{77.0$\pm$0.0} & \underline{75.0$\pm$0.0} & \textbf{77.0$\pm$0.0} & \textbf{78.0$\pm$0.0} & 74.0$\pm$0.0 & 63.0$\pm$0.0 & 70.0$\pm$0.0 & \textbf{77.0$\pm$0.0} \\
 & EGuR-3 & 41.0$\pm$0.0 & 32.0$\pm$0.0 & 28.0$\pm$0.0 & 40.0$\pm$0.0 & 65.0$\pm$0.0 & 43.0$\pm$0.0 & 68.0$\pm$0.0 & 64.0$\pm$0.0 & 60.0$\pm$0.0 & 44.0$\pm$0.0 & 38.0$\pm$0.0 & 37.0$\pm$0.0 \\
\bottomrule
\end{tabular}
\end{table}

\begin{table}[ht]
\centering
\tiny
\caption{Cost (\$) across training progress for different methods on 3-SAT and AIME datasets. Values shown as mean $\pm$ standard deviation across seeds. Claude uses EGuR-5, GPT and Qwen use EGuR-3. \textbf{Bold} indicates lowest cost within each model, \underline{underlined} indicates second lowest within each model.}
\label{tab:cost-1}
\begin{tabular}{llcccccccc}
\toprule
Model & Method & \multicolumn{4}{c}{3-SAT} & \multicolumn{4}{c}{AIME} \\
 &  & 0\% & 33\% & 66\% & 100\% & 0\% & 33\% & 66\% & 100\% \\
\midrule
\multirow{5}{*}{Claude-3.5-Sonnet} & CodeAct & 0.398$\pm$0.049 & 0.398$\pm$0.049 & 0.398$\pm$0.049 & 0.398$\pm$0.049 & 0.814$\pm$0.112 & 0.814$\pm$0.112 & 0.814$\pm$0.112 & 0.814$\pm$0.112 \\
 & EGuR-ZS & \underline{0.262$\pm$0.016} & \underline{0.262$\pm$0.016} & \underline{0.262$\pm$0.016} & \underline{0.262$\pm$0.016} & \underline{0.467$\pm$0.046} & \underline{0.467$\pm$0.046} & \underline{0.467$\pm$0.046} & \underline{0.467$\pm$0.046} \\
 & Mem0 & 0.386$\pm$0.029 & 0.371$\pm$0.040 & 0.349$\pm$0.059 & 0.396$\pm$0.011 & 0.880$\pm$0.075 & 0.718$\pm$0.107 & 0.837$\pm$0.091 & 0.733$\pm$0.124 \\
 & DC & 8.969$\pm$1.020 & 10.056$\pm$1.431 & 10.572$\pm$1.095 & 9.951$\pm$1.944 & 1.133$\pm$0.149 & 1.371$\pm$0.395 & 2.146$\pm$0.109 & 2.259$\pm$0.280 \\
 & EGuR-5 & \textbf{0.246$\pm$0.026} & \textbf{0.143$\pm$0.052} & \textbf{0.153$\pm$0.007} & \textbf{0.150$\pm$0.010} & \textbf{0.427$\pm$0.020} & \textbf{0.422$\pm$0.049} & \textbf{0.418$\pm$0.060} & \textbf{0.276$\pm$0.037} \\
\midrule
\multirow{5}{*}{GPT-OSS-120B} & CodeAct & 0.036$\pm$0.007 & 0.036$\pm$0.007 & 0.036$\pm$0.007 & 0.036$\pm$0.007 & \underline{0.004$\pm$0.000} & \textbf{0.004$\pm$0.000} & \underline{0.004$\pm$0.000} & \underline{0.004$\pm$0.000} \\
 & EGuR-ZS & \underline{0.005$\pm$0.001} & \underline{0.005$\pm$0.001} & \underline{0.005$\pm$0.001} & \underline{0.005$\pm$0.001} & 0.004$\pm$0.001 & \underline{0.004$\pm$0.001} & 0.004$\pm$0.001 & 0.004$\pm$0.001 \\
 & Mem0 & 0.027$\pm$0.002 & 0.031$\pm$0.005 & 0.038$\pm$0.009 & 0.035$\pm$0.004 & 0.006$\pm$0.002 & 0.010$\pm$0.001 & 0.007$\pm$0.001 & 0.011$\pm$0.003 \\
 & DC & 0.019$\pm$0.002 & 0.014$\pm$0.001 & 0.017$\pm$0.001 & 0.021$\pm$0.002 & 0.018$\pm$0.000 & 0.017$\pm$0.001 & 0.028$\pm$0.001 & 0.032$\pm$0.001 \\
 & EGuR-3 & \textbf{0.004$\pm$0.000} & \textbf{0.003$\pm$0.000} & \textbf{0.004$\pm$0.001} & \textbf{0.004$\pm$0.001} & \textbf{0.003$\pm$0.000} & 0.006$\pm$0.002 & \textbf{0.004$\pm$0.001} & \textbf{0.003$\pm$0.001} \\
\midrule
\multirow{5}{*}{Qwen3-Next-80B} & CodeAct & 0.251$\pm$0.024 & 0.251$\pm$0.024 & 0.251$\pm$0.024 & 0.251$\pm$0.024 & \underline{0.104$\pm$0.015} & \underline{0.104$\pm$0.015} & 0.104$\pm$0.015 & 0.104$\pm$0.015 \\
 & EGuR-ZS & \underline{0.137$\pm$0.013} & \underline{0.137$\pm$0.013} & \underline{0.137$\pm$0.013} & \underline{0.137$\pm$0.013} & 0.120$\pm$0.023 & 0.120$\pm$0.023 & 0.120$\pm$0.023 & 0.120$\pm$0.023 \\
 & Mem0 & 0.212$\pm$0.000 & 0.208$\pm$0.000 & 0.223$\pm$0.000 & 0.228$\pm$0.000 & 0.105$\pm$0.000 & 0.115$\pm$0.000 & \underline{0.080$\pm$0.000} & \underline{0.089$\pm$0.000} \\
 & EGuR-3 & \textbf{0.089$\pm$0.000} & \textbf{0.097$\pm$0.000} & \textbf{0.135$\pm$0.000} & \textbf{0.096$\pm$0.000} & \textbf{0.057$\pm$0.000} & \textbf{0.038$\pm$0.000} & \textbf{0.041$\pm$0.000} & \textbf{0.054$\pm$0.000} \\
\bottomrule
\end{tabular}
\end{table}

\begin{table}[ht]
\centering
\tiny
\caption{Cost (\$) across training progress for different methods on Movie Recommendation and Word Sorting datasets. Values shown as mean $\pm$ standard deviation across seeds. Claude uses EGuR-5, GPT and Qwen use EGuR-3. \textbf{Bold} indicates lowest cost within each model, \underline{underlined} indicates second lowest within each model.}
\label{tab:cost-2}
\begin{tabular}{llcccccccc}
\toprule
Model & Method & \multicolumn{4}{c}{Movie Rec.} & \multicolumn{4}{c}{Word Sort.} \\
 &  & 0\% & 33\% & 66\% & 100\% & 0\% & 33\% & 66\% & 100\% \\
\midrule
\multirow{5}{*}{Claude-3.5-Sonnet} & CodeAct & 0.142$\pm$0.004 & 0.142$\pm$0.004 & 0.142$\pm$0.004 & 0.142$\pm$0.004 & \textbf{0.060$\pm$0.007} & \textbf{0.060$\pm$0.007} & \underline{0.060$\pm$0.007} & \textbf{0.060$\pm$0.007} \\
 & EGuR-ZS & \underline{0.097$\pm$0.005} & \underline{0.097$\pm$0.005} & \underline{0.097$\pm$0.005} & \underline{0.097$\pm$0.005} & \underline{0.067$\pm$0.002} & \underline{0.067$\pm$0.002} & 0.067$\pm$0.002 & \underline{0.067$\pm$0.002} \\
 & Mem0 & 0.161$\pm$0.004 & 0.156$\pm$0.005 & 0.166$\pm$0.007 & 0.166$\pm$0.005 & 0.082$\pm$0.005 & 0.084$\pm$0.004 & 0.085$\pm$0.006 & 0.078$\pm$0.001 \\
 & DC & 0.826$\pm$0.011 & 2.525$\pm$0.047 & 2.710$\pm$0.034 & 2.885$\pm$0.014 & 1.214$\pm$0.047 & 2.822$\pm$0.133 & 5.426$\pm$0.066 & 4.322$\pm$0.068 \\
 & EGuR-5 & \textbf{0.087$\pm$0.001} & \textbf{0.066$\pm$0.002} & \textbf{0.061$\pm$0.001} & \textbf{0.049$\pm$0.001} & 0.077$\pm$0.003 & 0.071$\pm$0.001 & \textbf{0.059$\pm$0.000} & 0.072$\pm$0.005 \\
\midrule
\multirow{5}{*}{GPT-OSS-120B} & CodeAct & 0.003$\pm$0.000 & 0.003$\pm$0.000 & 0.003$\pm$0.000 & \underline{0.003$\pm$0.000} & 0.003$\pm$0.000 & 0.003$\pm$0.000 & 0.003$\pm$0.000 & 0.003$\pm$0.000 \\
 & EGuR-ZS & \underline{0.002$\pm$0.000} & \underline{0.002$\pm$0.000} & \underline{0.002$\pm$0.000} & \textbf{0.002$\pm$0.000} & \textbf{0.001$\pm$0.000} & \textbf{0.001$\pm$0.000} & \textbf{0.001$\pm$0.000} & \underline{0.001$\pm$0.000} \\
 & Mem0 & 0.003$\pm$0.000 & 0.003$\pm$0.000 & 0.003$\pm$0.000 & 0.003$\pm$0.000 & 0.004$\pm$0.000 & 0.004$\pm$0.000 & 0.004$\pm$0.000 & 0.004$\pm$0.000 \\
 & DC & 0.038$\pm$0.001 & 0.093$\pm$0.001 & 0.099$\pm$0.000 & 0.069$\pm$0.001 & 0.038$\pm$0.001 & 0.059$\pm$0.001 & 0.082$\pm$0.002 & 0.052$\pm$0.001 \\
 & EGuR-3 & \textbf{0.001$\pm$0.000} & \textbf{0.001$\pm$0.000} & \textbf{0.001$\pm$0.000} & 0.005$\pm$0.000 & \underline{0.001$\pm$0.000} & \underline{0.002$\pm$0.000} & \underline{0.001$\pm$0.000} & \textbf{0.001$\pm$0.000} \\
\midrule
\multirow{5}{*}{Qwen3-Next-80B} & CodeAct & 0.096$\pm$0.003 & 0.096$\pm$0.003 & 0.096$\pm$0.003 & 0.096$\pm$0.003 & 0.083$\pm$0.003 & 0.083$\pm$0.003 & 0.083$\pm$0.003 & 0.083$\pm$0.003 \\
 & EGuR-ZS & 0.073$\pm$0.005 & 0.073$\pm$0.005 & 0.073$\pm$0.005 & 0.073$\pm$0.005 & \underline{0.078$\pm$0.009} & \underline{0.078$\pm$0.009} & \underline{0.078$\pm$0.009} & \textbf{0.078$\pm$0.009} \\
 & Mem0 & \underline{0.071$\pm$0.000} & \underline{0.066$\pm$0.000} & \underline{0.067$\pm$0.000} & \underline{0.063$\pm$0.000} & 0.100$\pm$0.000 & 0.118$\pm$0.000 & 0.102$\pm$0.000 & 0.110$\pm$0.000 \\
 & EGuR-3 & \textbf{0.047$\pm$0.000} & \textbf{0.049$\pm$0.000} & \textbf{0.043$\pm$0.000} & \textbf{0.056$\pm$0.000} & \textbf{0.068$\pm$0.000} & \textbf{0.052$\pm$0.000} & \textbf{0.053$\pm$0.000} & \underline{0.083$\pm$0.000} \\
\bottomrule
\end{tabular}
\end{table}

\begin{table}[ht]
\centering
\tiny
\caption{Cost (\$) across training progress for different methods on Object Counting dataset. Values shown as mean $\pm$ standard deviation across seeds. Claude uses EGuR-5, GPT and Qwen use EGuR-3. \textbf{Bold} indicates lowest cost within each model, \underline{underlined} indicates second lowest within each model.}
\label{tab:cost-3}
\begin{tabular}{llcccc}
\toprule
Model & Method & \multicolumn{4}{c}{Object Count.} \\
 &  & 0\% & 33\% & 66\% & 100\% \\
\midrule
\multirow{5}{*}{Claude-3.5-Sonnet} & CodeAct & \textbf{0.095$\pm$0.005} & \underline{0.095$\pm$0.005} & \underline{0.095$\pm$0.005} & \underline{0.095$\pm$0.005} \\
 & EGuR-ZS & 0.234$\pm$0.020 & 0.234$\pm$0.020 & 0.234$\pm$0.020 & 0.234$\pm$0.020 \\
 & Mem0 & \underline{0.125$\pm$0.001} & 0.127$\pm$0.005 & 0.125$\pm$0.005 & 0.128$\pm$0.009 \\
 & DC & 1.539$\pm$0.046 & 8.155$\pm$0.295 & 8.803$\pm$0.931 & 7.156$\pm$0.397 \\
 & EGuR-5 & 0.242$\pm$0.003 & \textbf{0.060$\pm$0.001} & \textbf{0.063$\pm$0.001} & \textbf{0.064$\pm$0.001} \\
\midrule
\multirow{5}{*}{GPT-OSS-120B} & CodeAct & 0.020$\pm$0.001 & 0.020$\pm$0.001 & 0.020$\pm$0.001 & 0.020$\pm$0.001 \\
 & EGuR-ZS & \underline{0.006$\pm$0.001} & \underline{0.006$\pm$0.001} & \textbf{0.006$\pm$0.001} & \underline{0.006$\pm$0.001} \\
 & Mem0 & 0.020$\pm$0.003 & 0.022$\pm$0.001 & 0.021$\pm$0.003 & 0.019$\pm$0.000 \\
 & DC & 0.200$\pm$0.009 & 0.250$\pm$0.019 & 0.194$\pm$0.010 & 0.128$\pm$0.048 \\
 & EGuR-3 & \textbf{0.005$\pm$0.000} & \textbf{0.004$\pm$0.000} & \underline{0.006$\pm$0.000} & \textbf{0.006$\pm$0.001} \\
\midrule
\multirow{5}{*}{Qwen3-Next-80B} & CodeAct & \underline{0.049$\pm$0.001} & \underline{0.049$\pm$0.001} & \underline{0.049$\pm$0.001} & \underline{0.049$\pm$0.001} \\
 & EGuR-ZS & 0.086$\pm$0.010 & 0.086$\pm$0.010 & 0.086$\pm$0.010 & 0.086$\pm$0.010 \\
 & Mem0 & 0.055$\pm$0.000 & 0.057$\pm$0.000 & 0.064$\pm$0.000 & 0.065$\pm$0.000 \\
 & EGuR-3 & \textbf{0.049$\pm$0.000} & \textbf{0.032$\pm$0.000} & \textbf{0.029$\pm$0.000} & \textbf{0.025$\pm$0.000} \\
\bottomrule
\end{tabular}
\end{table}

\section{Example Generated Strategies}
\label{app:strat-qual}
The following shows an example of a strategy generated with GPT-OSS-120B to solve a sarcasm task (Sarc Triples) from BBEH. We can see that this strategy includes a prompt tailored to the specific problem, majority voting, and modification of the temperature parameter to the called model.

\begin{promptblock}[lang=Python]{Generated Strategy}
def strategy(x):
  """
  Detect sarcasm in three Reddit replies.
  Repeatedly query the model, extract the list after "FINAL ANSWER:",
  and return the majority-voted label string.
  """
  import re
  from collections import Counter

  # Prompt that enforces step-by-step reasoning and a strict final answer format
  base_prompt = (
  "You are given three (post, reply) pairs. For each reply decide whether it is "
  "sarcastic (1) or not sarcastic (0). Think carefully about each pair, then output "
  "the three labels in order as a comma-separated list. Write the list after a line "
  "that starts exactly with 'FINAL ANSWER:'.\n"
  f"Problem: {x}"
  )

  collected_answers = []

  # Number of independent attempts (adjustable; 5 balances cost and reliability)
  for _ in range(5):
    # First generation
    out = M(base_prompt, temperature=0.5, max_tokens=2000)[0]

    # Try to extract the answer
    ans = extract_final_answer(out)
    if not ans:
      # Ask the model to repeat with the required format
      clarification = (
        "Please respond again, making sure to include a line that starts exactly with "
        "'FINAL ANSWER:' followed by the three labels separated by commas."
      )
      out = M(clarification + "\nPrevious output:\n" + out,
              temperature=0, max_tokens=2000)[0]
      ans = extract_final_answer(out)

    if ans:
      # Normalise whitespace and store
      collected_answers.append(ans.strip())

  if not collected_answers:
    return "No answer found."

  # Majority vote over the collected answer strings
  most_common = Counter(collected_answers).most_common(1)[0][0]
  return most_common
\end{promptblock}

For Claude 3.7 Sonnet, we show generated strategies before and after training on the BBEH Object Counting task. The first strategy below, from before training answers the particular problem it was generated for incorrectly and cost \$0.166 while the following strategy answers the same problem correctly and only costs \$0.062.

\begin{promptblock}[lang=Python]{BBEH Object Counting strategy from Claude 3.7 Sonnet before training}
async def strategy(x):
    SYSTEM_PROMPT = """You are an expert at solving counting and categorization
problems with Python code.

Task: Solve the question by iteratively writing and executing Python code.

Rules:
- Reply with code inside a single ```python ...``` block. The code will be executed
  for you; anything you print will be returned to you next round.
- Read the execution result, and then decide on how to proceed next. You must iterate
  until you have solved the problem.
- When ready, return the solution exactly as: FINAL ANSWER: <answer>
  (You may print this from code or write it directly outside the code block.)

Constraints:
- Code must be self-contained, use only the Python standard library, and avoid
  network/file I/O and pip installs.

Approach:
1. Parse the text to identify all items the narrator ("I") has
2. Categorize each item as either a musical instrument or an animal/insect
3. Count the total number of each category
4. Calculate the absolute difference between the two totals"""

    PARSE_ERROR_PROMPT = "Parsing error: I couldn't find a ```python ... ``` code \
    block or a line starting with 'FINAL ANSWER:' in the model output. \
    Please either:\n1) Provide valid Python within ```python ... ``` OR\n2) \
    Reply directly with a line exactly: FINAL ANSWER: <answer> with no trailing text."

    max_rounds = 20
    messages = [{"role": "system", "content": SYSTEM_PROMPT}]

    messages.append({"role": "user", "content": f"Problem: {x}"})
    out = await model_call(messages, temperature=0.7, thinking=True, max_tokens=20000)
    messages.append({"role": "assistant", "content": out})
    execution_state = {}
    for i in range(max_rounds):
        # First check for any output code to execute
        blocks = extract_python_code(out)
        if blocks:
            block = "\n\n".join(blocks)
            exec_res, execution_state = python_exec(block, execution_state)

            # Check for final answer in execution output
            m = extract_final_answer(exec_res)
            if m:
                return m.strip()

            # No final answer, so take another turn of model output
            messages.append({
                "role": "user",
                "content": f"The output of executing the generated Python code \
                    is:\n{exec_res}"})
            out = await model_call(
                messages, temperature=0.7, thinking=True, max_tokens=20000)
            messages.append({"role": "assistant", "content": out})
            continue
        
        # If no code, then check for a final answer
        m = extract_final_answer(out)
        if m:
            return m.strip()
        messages.append({"role": "user", "content": PARSE_ERROR_PROMPT})
        out = await model_call(messages, temperature=0.7, thinking=True, max_tokens=20000)
        messages.append({"role": "assistant", "content": out})

    messages.append({
        "role": "user",
        "content": "Now output the final answer on a single line as \
                   'FINAL ANSWER: answer' with no trailing text."})
    out = await model_call(messages, temperature=0.7, thinking=True, max_tokens=20000)
    m = extract_final_answer(out)
    if m:
        return m.strip()
    else:
        raise Exception("No answer found.")
\end{promptblock}

\begin{promptblock}[lang=Python]{BBEH Object Counting strategy from Claude 3.7 Sonnet after training}
async def strategy(x):
    PROMPT = """Solve the following problem step by step:

1. First, carefully read through the text and identify all items that "I" 
   (the narrator) have. Ignore items that belong to other people
   (grandmother, friend, sister, brother, father, mother, aunt, grandfather,
   cousin, uncle).
2. For each item that belongs to "I", categorize it as either:
   - An animal/insect (e.g., ducks, rabbits, walrus, hens, moose, wasps, bees,
     chameleons, newts, vultures, otter, tigers, shrimps, elks, baboons, yaks,
     boa snakes, anteaters, magpies, fish, narwhals, chimpanzees, antelopes)
   - A musical instrument (e.g., harps, panpipes, balafons, marimbas,
     electric guitars, recorders, clavichord, electric pianos, tanpuras, ouds,
     bass guitars)
   - Other items (cars like nissan pathfinder, kia niro, honda accord,
     nissan frintier, toyota highlander, hyundai elantra, mazda2, buick enclave,
     hyundai santa fe) which we will ignore for this problem
3. Create a table with columns: Item, Category (Animal/Insect, Musical Instrument,
   Other), Quantity.
4. For items with stories about changing quantities, ALWAYS use the FINAL quantity
   mentioned. For example:
   - "initially I had X, then I lost Y, but then I got Z, but then I lost W,
      so at the end I am left with V" → Count as V
   - "initially I had X, then I also got Y making it a total of Z, then I got W
       making it a total of A, but then I lost B of them so eventually I am left
       with C" → Count as C
5. Count the total number of animals/insects that "I" have.
6. Count the total number of musical instruments that "I" have.
7. Calculate the absolute difference between these two totals.

Be methodical and organize your work clearly. After completing your analysis,
write the final answer on a single line as 'FINAL ANSWER: <answer>' with no trailing
text.

Problem: {x}"""

    MISSING_ANSWER_FEEDBACK = "Your last response did not include a line exactly of \
        the form 'FINAL ANSWER: <answer>'. Please try again to output a final answer,\
        responding with exactly one line 'FINAL ANSWER: <answer>'."

    messages = [{"role": "user", "content": PROMPT.format(x=x)}]
    for _ in range(5):
        output = await model_call(
            messages, temperature=0.7, thinking=True, max_tokens=20000)
        # Find FINAL ANSWER
        ans = extract_final_answer(output)
        if ans:
            return ans
        messages.append({"role": "assistant", "content": output})
        messages.append({"role": "user", "content": MISSING_ANSWER_FEEDBACK})
    
    # One last attempt with a more direct prompt
    messages.append({
        "role": "user",
        "content": "Based on your analysis, what is the absolute difference between \
        the total number of animals/insects and musical instruments that the narrator \
        has? Provide only the numerical answer in the format 'FINAL ANSWER: <number>'."})
    output = await model_call(
        messages, temperature=0.7, thinking=True, max_tokens=20000)
    ans = extract_final_answer(output)
    if ans:
        return ans
    else:
        raise Exception("No answer found after multiple attempts")
\end{promptblock}

\section{Example Learned Context}
\label{app:qual-context}

The final memory resulting from \ourabbrev-5 with Claude 3.7 Sonnet on BBEH Movie Recommendations is shown below. The memory consists of several memory entries and a single entry containing the (task, best strategy) pair which is used for solving future problems.

\begin{promptblock}{Learned Context}
<memory_entry-1>
When analyzing similarity between items for group appeal (movies, books, etc.), consider these key factors:
1. Era/time period consistency (items from the same era tend to have more similar appeal)
2. Language/accessibility (foreign language items create significant outliers)
3. Target demographics (mainstream vs. niche appeal)
4. Critical reception and popular appeal
5. Production values and stylistic approaches

For movie similarity specifically, the most reliable indicators of similar group appeal are:
- Same language and era
- Similar accessibility factors
- Comparable mainstream vs. niche positioning
- Consistent tone and pacing
</memory_entry-1>

<memory_entry-2>
For similarity analysis problems, the Chain of Thought (CoT) approach with deep thinking tends to be more cost-effective than more complex strategies like CodeAct or Evaluator-Optimizer when the task involves subjective judgment rather than algorithmic calculation.

The CoT approach with a well-structured prompt that guides the analysis through specific criteria (genres, audience, reception, etc.) and asks for explicit rating/ranking of options provides sufficient structure for the model to make accurate comparisons without requiring the overhead of code execution or multiple model calls.
</memory_entry-2>

[...]

<memory_entry-5>
Task: Analyzing which group of movies has the most similar audience appeal
Best Strategy:
<strategy>
async def strategy(x):
    PROMPT = f"""You are an expert film critic tasked with analyzing which group of movies would have the most similar appeal to the same audience. Think step by step and provide your final answer on a single line as 'FINAL ANSWER: <option letter>' with no trailing text.

When analyzing similarity between movies for group appeal, consider these key factors:
1. Era/time period consistency (movies from the same era tend to have more similar appeal)
2. Language/accessibility (foreign language films create significant outliers)
3. Target demographics (mainstream vs. niche appeal)
4. Critical reception and popular appeal
5. Production values and stylistic approaches
6. Psychological appeal patterns (puzzle-solving vs. emotional engagement vs. escapism)
7. Viewing experience requirements (active vs. passive engagement)
8. Narrative complexity and accessibility
9. Emotional responses elicited (fear, humor, intellectual stimulation, etc.)

For movie similarity specifically, the most reliable indicators of similar group appeal are:
- Same language and era
- Similar accessibility factors
- Comparable mainstream vs. niche positioning
- Consistent tone and pacing

Significant outliers that reduce group cohesion include:
- Animated shorts among feature films
- Family/children's content mixed with adult-oriented films
- Critically panned films grouped with acclaimed ones
- Comedic/light entertainment mixed with serious dramas
- Foreign language films mixed with English-language films
- Adult/erotic content mixed with general audience films

For each option:
1. Analyze each movie in the group individually
2. Compare each movie against every other movie in the group
3. Identify specific outliers that would reduce group cohesion
4. Rate the overall cohesion on a 1-10 scale
5. Explain your reasoning thoroughly

Problem: {x}

Think through each option carefully and systematically.
"""
    MISSING_ANSWER_FEEDBACK = "Your last response did not include a line exactly of the form 'FINAL ANSWER: <option letter>'. Please try again to output a final answer, responding with exactly one line 'FINAL ANSWER: <option letter>'."

    messages = [{"role": "user", "content": PROMPT}]
    for _ in range(5):
        output = await model_call(messages, temperature=1.0, thinking=True, max_tokens=20000)
        # Find FINAL ANSWER
        ans = extract_final_answer(output)
        if ans:
            return ans
        messages.append({"role": "assistant", "content": output})
        messages.append({"role": "user", "content": MISSING_ANSWER_FEEDBACK})
    raise Exception("No answer found")
</strategy>
</memory_entry-5>
\end{promptblock}

\section{Prompts}
\label{app:prompts}

We include the complete prompts for the \guide and \updater in the following sections. These prompts define the two strategies defining \ourabbrev.

\subsection{\guide}
\label{app:consultant}

The prompt for the \guide is included below.

\includeprompt{\guide Prompt}{prompts/consultant.md}

\subsection{\updater}
\label{app:contextupdater}

The prompt for the \updater is included below.
\includeprompt{\updater Prompt}{prompts/memory-updater.md}

\end{document}